\PassOptionsToPackage{table}{xcolor}
\pdfoutput=1

\documentclass[11pt]{article}

\usepackage[final]{acl}

\usepackage{times}
\usepackage{latexsym}
\usepackage{hyperref}
\usepackage{url}
\usepackage{booktabs}
\usepackage{graphicx}
\usepackage{subcaption}
\usepackage{longtable}
\usepackage{ltablex}
\keepXColumns
\usepackage{adjustbox}
\usepackage{multirow,multicol}
\usepackage{cleveref}
\usepackage{placeins}

\usepackage[T1]{fontenc}

\usepackage[utf8]{inputenc}

\usepackage{microtype}

\usepackage{inconsolata}

\newcommand{\xlinstruct}{\texttt{XL-Instruct}}
\newcommand{\xlalpacaeval}{\texttt{XL-AlpacaEval}}
\newcommand{\malpacaeval}{\texttt{m-AlpacaEval}}
\newcommand{\xlsuite}{\texttt{XL-Suite}}

\title{XL-Suite: Cross-Lingual Synthetic Training and Evaluation Data for Open-Ended Generation}

\author{%
Vivek Iyer$^{1}$\qquad
Pinzhen Chen$^{1,3}$\qquad
Ricardo Rei$^{2,}$\thanks{Work done while at Unbabel.}\qquad
Alexandra Birch$^{1}$ \\
$^{1}$University of Edinburgh\qquad
$^{2}$Sword Health\qquad
$^{3}$Queen's University Belfast \\
\texttt{vivek.iyer@ed.ac.uk}
}

\begin{document}
\maketitle
\begin{abstract}
Cross-lingual open-ended generation---responding in a language different from that of the query---is an important yet understudied problem. This work proposes \xlinstruct{}, a novel technique for generating high-quality synthetic data. We also introduce \xlalpacaeval{}, a new benchmark for evaluating cross-lingual generation capabilities of large language models (LLMs). Our experiments show that fine-tuning with just 8K instructions generated using our \xlinstruct{} significantly improves model performance: increasing the win rate against GPT-4o-mini from 7.4\% to 21.5\% and improving on several fine-grained quality metrics. Moreover, base LLMs fine-tuned on \xlinstruct{} exhibit strong zero-shot improvements to same-language question answering, as shown on our machine-translated \malpacaeval{}. These consistent gains highlight the promising role of \xlinstruct{} in the post-training of multilingual LLMs. Finally, we publicly release \xlsuite{}, a collection of training and evaluation data to facilitate research in cross-lingual open-ended generation.

\end{abstract}

\section{Introduction}

Cross-lingual generation is the task of understanding a query in a given source language and generating a response in a different target language. This task has assumed greater relevance in the recent era of large language models (LLMs) with multilingual capabilities.
\citet{marchisio-etal-2024-understanding} noted its usefulness for a) companies that serve such LLMs across dozens of languages, but \textit{optimizing a prompt for each input language
is inefficient in practice}, and b) \textit{when a user needs a generation in a language they do not speak}. The conventional cascaded approaches to cross-lingual generation \citep{huang-etal-2023-languages, qin-etal-2023-cross, li2024think} could be problematic due to the noisy nature of machine translation, which leads to information loss or an unnatural-sounding response. It is also wasteful of inference time and cost, since the intermediary English response is thrown away once the desired cross-lingual output is obtained.

\begin{figure*}[t]
    \centering
        \includegraphics[width=0.9\textwidth]{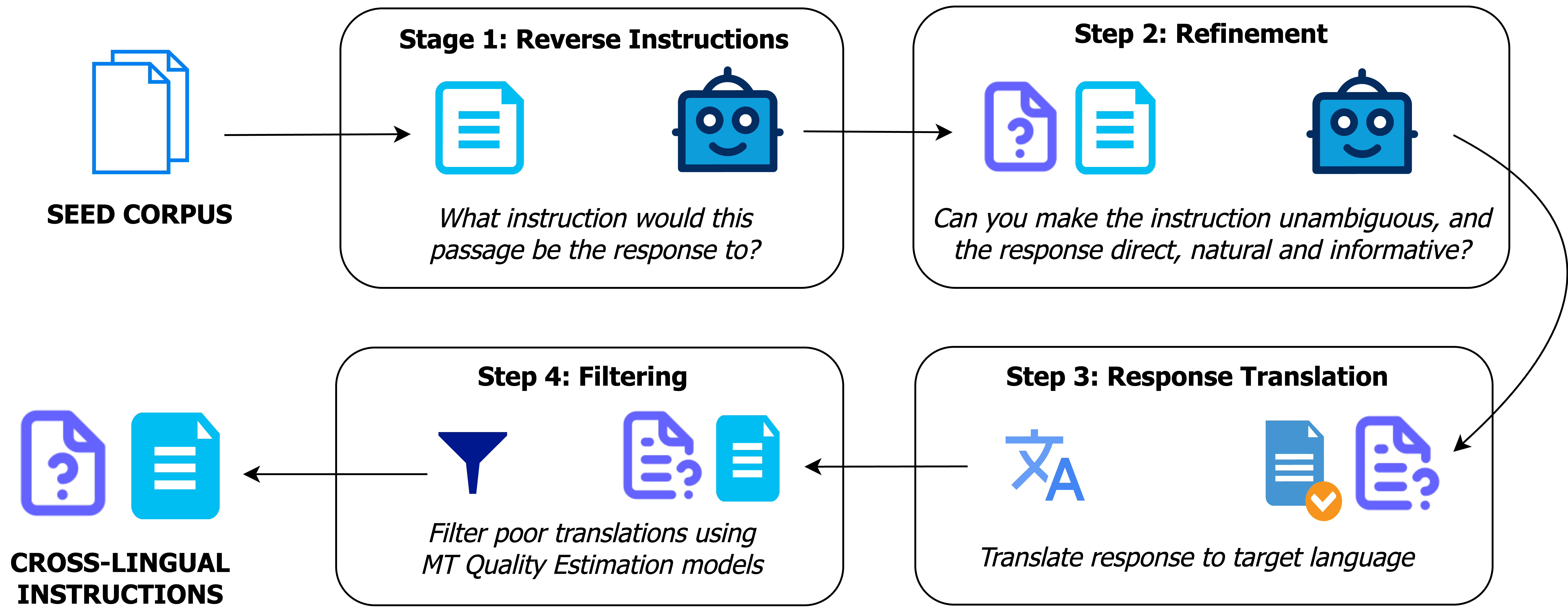}
    \caption{The \xlinstruct{} pipeline: 1) instruction generation from seed English data; 2) data refinement; 3) response translation into non-English; 4) data filtering, with more details in \Cref{sec:xl-instruct}.}
    \label{fig:xl-instruct} 
\end{figure*}

The adaptation of LLMs to cross-lingual open-ended generation is relevant, given their versatile capabilities in both language conversion and question answering, but remains understudied. A primary reason is the absence of high-quality datasets and evaluation benchmarks. This work addresses the data deficiency for the cross-lingual generation task from both the modelling and evaluation perspectives. We first introduce \textbf{\xlalpacaeval{}}, a cross-lingual evaluation benchmark built on AlpacaEval \citep{alpaca_eval}, and we observe poor off-the-shelf performance for most open-source multilingual LLMs. As a solution, we propose \textbf{\xlinstruct{}}, a synthetic data generation technique to create high-quality cross-lingual data at scale (illustrated in Figure \ref{fig:xl-instruct}) and show that fine-tuning with \xlinstruct{} significantly and consistently boosts cross-lingual performance across a range of base and instruction-tuned LLMs. 
Beyond cross-lingual capabilities, we also created a machine-translated benchmark for same-language generation, \textbf{\malpacaeval{}}, to demonstrate that our proposed data synthesis method achieves strong zero-shot transfer performance.

In this work, we seek to answer the following research questions through a comprehensive set of experiments:
\begin{itemize}
    \item \textbf{RQ1:} How good are off-the-shelf multilingual LLMs in cross-lingual generation? (§\ref{sec:x-alpacaeval})
    \item \textbf{RQ2:} How does \xlinstruct{} improve cross-lingual capabilities of various LLMs? (§\ref{sec:experiments_xl})
    \item \textbf{RQ3:} How does \xlinstruct{} fine-tuning impact standard multilingual same-language question answering? (§\ref{sec:experiments_m})
\end{itemize}

Finally, to facilitate research in the cross-lingual LLM domain, which currently lacks sufficient resources for both evaluation and post-training, we publicly release \xlsuite{}\footnote{\url{https://huggingface.co/collections/viyer98/xl-suite-68ceb97cb1cc7e8499ffb971}} -- a comprehensive collection of cross-lingual training (\xlinstruct{}) and evaluation (\xlalpacaeval{} and \malpacaeval{}) data.
\label{sec:introduction}

\section{Related Work}
\paragraph{Cross-Lingual LLM Prompting} Most of the current research on cross-lingual generation in LLMs focuses on prompting strategies. The primary goal of generation here is to leverage the extensive knowledge and superior reasoning capabilities of LLMs in high-resourced languages (like English) to improve the final answer in lower-resourced ones \citep{qin-etal-2023-cross, huang-etal-2023-languages, singh2024three, wang-etal-2025-large}. Similarly motivated, PLUG \citep{zhang-etal-2024-plug} fine-tunes an LLM for this cross-lingual process: it first answers a non-English question by reasoning in English, then translates the response to the target language. Other extensions to this cross-lingual prompting paradigm have also emerged, such as X-InSTA \citep{tanwar-etal-2023-multilingual}, which uses a semantic encoder to select relevant cross-lingual examples, while SITR \citep{li2024think} employs self-reflection and iterative refinement to improve cross-lingual summarization. However, no prior study has approached cross-lingual open-ended generation as the 
primary training objective.

\paragraph{Data Synthesis} Previous studies on the creation of synthetic data for post-training LLMs have mostly been 
limited to monolingual scenarios, mostly in English. 
Self-Instruct \citep{wang-etal-2023-self-instruct} and Unnatural Instructions \citep{honovich-etal-2023-unnatural} were among the first to show how LLMs could be used to generate instructions from seed data. 
Later efforts have focused on generating diversified and skill-specific synthetic data. Tülu 3 \citep{lambert2024t}, for instance, used persona-driven prompting to yield diverse synthetic instructions \citep{ge2024scaling}, while Llama 3 \citep{dubey2024llama} leveraged skill-specific experts as teacher models to generate data for coding, math, multilinguality, etc.
To enable multilingual support, machine translation is often used to extend English resources to other languages \citep{muennighoff-etal-2023-crosslingual,lai-etal-2023-okapi,ranaldi-pucci-2023-english,chen-etal-2024-monolingual}. 
Given that English resources are often model outputs (e.g., of ChatGPT), training on translations of these can limit models' exposure to diversity.

\paragraph{Reverse Instruction} A subset of data synthesis approaches relevant to our work is called ``reverse instruction'', which generates instructions from seed data and then uses the original seed data as responses to these instructions, with back-translation being an early prominent example \citep{sennrich-etal-2016-improving}. Our work follows this trend of approaches applied to LLMs, where initial works \citep{li2023self, wang2023harnessing} presented a two-step procedure that can be done iteratively: 1) fine-tuning a model to perform instruction generation, followed by 2) heuristic-based filtering to keep high-quality synthetic data. Later, \citet{chen2023dog} proposed ``instruction wrapping'' to refine response quality before fine-tuning the reverse instruction model.  LongForm \citep{koksal2023longform} bypassed the fine-tuning step and leveraged a strong ``teacher'' LLM (InstructGPT) to generate such instructions directly, yielding significant improvements in English text generation tasks. 
MURI \citep{koksal2024muri} and X-Instruction \citep{li-etal-2024-x} extend LongForm to multilingual generation.  The former back-translates to English, generates reverse instructions, and then forward-translates to low-resource languages. 
The latter bypasses back-translation to English and queries the teacher LLM in the low-resource language directly, potentially exposing the synthetic data to quality issues. The focus of these works is on improving same-language generation performance. Finally, \citet{iyer-etal-2024-quality} and \citet{iyer-etal-2024-exploring} use similar strategies to create low-resource cross-lingual data for boosting MT performance of LLMs.

Unlike these previous works, our primary goal is to contribute data resources for cross-lingual open-ended generation, which includes a synthetic dataset where the instruction and response are in different languages, as well as a cross-lingual evaluation benchmark.\footnote{To the best of our knowledge, we are the first to propose a cross-lingual open-ended generation benchmark, and our synthetic training dataset is among the few publicly available.} 
Our experiments (see Table \ref{tab:xli_vs_xins}) show that it is of much higher quality than the closest prior work, X-Instruction \citep{li-etal-2024-x}.
We intend to release the \xlinstruct{} dataset under a permissive open source license.
\label{sec:related-work}

\section{\xlalpacaeval{}: A Cross-Lingual Evaluation Benchmark}

\paragraph{Dataset} To evaluate cross-lingual open-ended generation, we create the \textbf{\xlalpacaeval{}} benchmark, which is adapted from AlpacaEval v1 \citep{alpaca_eval}. AlpacaEval contains 805 multi-domain prompts sampled from various test sets \citep{dubois2024alpacafarm}, including OpenAssistant \citep{kopf2024openassistant}, Koala \citep{koala_blogpost_2023}, Vicuna \citep{chiang2023vicuna}, Self-Instruct \citep{wang-etal-2023-self-instruct} and Anthropic's Helpfulness test set \citep{bai2022training}. Evaluation is carried out through the LLM-as-a-judge approach \citep{zheng2023judging}, where an evaluator LLM is used to estimate how often a model output would be preferred by humans over a baseline reference.

To create \xlalpacaeval{}, we first manually examine the AlpacaEval dataset and filter out prompts that are tailored towards eliciting responses in English. For example, questions about correcting grammar in an English sentence cannot be answered cross-lingually (refer to \Cref{sec:manual_verification} for a detailed justification and a list of excluded prompts). The filtered test set consists of 797 prompts. Next, we add cross-lingual generation instructions (such as ``Answer in \texttt{\{language\}}'') to prompts randomly sampled from a list of templates (in \Cref{sec:generationprompts_appendix}) and create an evaluation set for eight languages, spanning resource availability, writing script, and geographical location: German (\texttt{deu}), Portuguese (\texttt{por}), Hungarian (\texttt{hun}), Lithuanian (\texttt{lit}), Irish (\texttt{gle}), Maltese (\texttt{mlt}), simplified Chinese (\texttt{zho}), and Hindi (\texttt{hin}). We focus on the En-X direction in this work, as generating in non-English is usually more challenging for LLMs that are usually English-centric. It should be straightforward to extend our benchmark to other languages and pairs---by appending the cross-lingual templated instructions to our filtered test set.

\paragraph{Evaluation} While the original implementation used GPT-4-turbo as both reference and evaluator models, we use GPT-4o-mini for reference and GPT-4o as the judge, given GPT-4o's strong multilingual capabilities. Our choice of using GPT-4o-mini as the reference model is motivated by two reasons: 1) we experiment with \textasciitilde7--9B LLMs in this work, making the GPT-4o-mini model a suitable baseline; and 2) using different reference and judge models, with the more capable one as the judge, should mitigate self-preference bias of models \citep{wataoka2024self}. Finally, GPT-4o has also been shown to obtain state-of-the-art pairwise correlations with human ratings in multilingual chat scenarios \citep{gureja2024m, son2024mm}.
\begin{table*}[thp]
\centering\small
\begin{tabular}{lrrrrrrrrrr}
\toprule
\multirow[b]{2}{*}{\textbf{Model}} & \multirow[b]{2}{*}{\textbf{Avg}} & \multicolumn{2}{c}{\textbf{High-Res EU}} & \multicolumn{2}{c}{\textbf{Med-Res EU}} & \multicolumn{2}{c}{\textbf{Low-Res EU}} & \multicolumn{2}{c}{\textbf{Non-EU}} \\
\cmidrule(lr){3-4} \cmidrule(lr){5-6} \cmidrule(lr){7-8} \cmidrule(lr){9-10}
 &  & \textbf{por} & \textbf{deu} & \textbf{hun} & \textbf{lit} & \textbf{gle} & \textbf{mlt} & \textbf{zho} & \textbf{hin} \\
\midrule
Salamandra 7B Instruct & 6.44 & 8.64 & 8.27 & 5.08 & 9.51 & 5.63 & 4.95 & 5.24 & 4.23 \\
Aya 23 8B & 8.85 & 17.04 & 15.04 & 2.07 & 2.22 & 2.45 & 1.92 & 9.46 & 20.57 \\
EuroLLM 9B Instruct & 12.70 & 18.94 & 16.49 & 8.66 & 16.57 & 9.37 & 8.51 & 14.82 & 8.23 \\
Qwen 2.5 7B Instruct & 16.73 & 30.88 & 16.35 & 6.82 & 14.68 & 7.17 & 3.69 & 44.63 & 9.59 \\
Gemma 2 9B IT & 23.29 & 35.42 & 32.08 & 19.80 & \textbf{27.28 }& 10.09 & \textbf{10.03} & 28.12 & 23.50 \\
Llama 3.1 8B Instruct & 24.36 & 40.28 & 35.72 & \textbf{23.07} & 20.74 & \textbf{13.20} & 8.47 & 31.21 & 22.22 \\
Aya Expanse 8B & \textbf{35.67} & \textbf{62.75} & \textbf{60.27} & 8.62 & 19.54 & 10.43 & 9.51 & \textbf{57.22} & \textbf{56.99} \\
\bottomrule
\end{tabular}%
\caption{Zero-shot win rates against GPT-4o-mini on \xlalpacaeval{} as judged by GPT-4o.}
\label{tab:xalpacaeval_zsresults}
\end{table*}

\begin{table*}[thp]
  \centering
  \small
  \begin{tabular}{lrrrrrrrrr}
    \toprule
    Model & \textbf{Avg} & \textbf{por} & \textbf{deu} & \textbf{hun} & \textbf{lit} & \textbf{gle} & \textbf{mlt} & \textbf{zho} & \textbf{hin} \\
    \midrule
    Salamandra 7B Instruct   & \cellcolor[gray]{0.88}4.45 & \cellcolor[gray]{0.89}3.32 & \cellcolor[gray]{0.89}2.47 & \cellcolor[gray]{0.89}2.16 & \cellcolor[gray]{0.89}3.71 & \cellcolor[gray]{0.86}7.49 & \cellcolor[gray]{0.86}8.00 & \cellcolor[gray]{0.87}6.09 & \cellcolor[gray]{0.89}2.37 \\
    Aya 23 8B               & \cellcolor[gray]{0.90}1.28 & \cellcolor[gray]{0.90}1.12 & -1.78  & -8.62  & \cellcolor[gray]{0.88}4.58 & \cellcolor[gray]{0.88}4.52 & \cellcolor[gray]{0.84}11.86 & \cellcolor[gray]{0.89}3.11 & -4.59 \\
    EuroLLM 9B Instruct      & \cellcolor[gray]{0.88}5.26 & -1.57 & -0.83   & \cellcolor[gray]{0.88}5.50 & \cellcolor[gray]{0.88}5.14 & \cellcolor[gray]{0.80}18.66 & \cellcolor[gray]{0.87}6.83 & \cellcolor[gray]{0.84}11.85 & -3.54 \\
    Qwen 2.5 7B Instruct     & -1.25  & -20.01 & \cellcolor[gray]{0.89}2.92 & \cellcolor[gray]{0.90}1.59 & \cellcolor[gray]{0.89}2.91 & \cellcolor[gray]{0.89}3.62 & \cellcolor[gray]{0.88}4.24 & \cellcolor[gray]{0.88}3.80 & -9.08 \\
    Gemma 2 9B IT            & -4.73  & -11.00 & -12.66 & -4.10  & -2.58  & \cellcolor[gray]{0.89}2.67 & -0.37  & \cellcolor[gray]{0.87}6.54 & -16.37 \\
    Llama 3.1 8B Instruct    & -10.55 & -23.84 & -18.01 & -7.96  & -8.14  & -1.75 & -2.69 & -0.08 & -21.92 \\
    Aya Expanse 8B           & -20.53 & -39.60 & -39.25 & -36.13 & -0.51  & -1.18 & -2.50 & -1.78 & -43.29 \\
    \bottomrule
  \end{tabular}
  \caption{Performance change over zero-shot when using Reason-then-Translate: scores represent differences against win rates from \Cref{tab:xalpacaeval_zsresults}. Strong positive improvements are shaded.}
  \label{tab:rtt-results}
\end{table*}

\paragraph{Models} To evaluate off-the-shelf cross-lingual capabilities of existing multilingual LLMs, we benchmark several strong open-weight models in the \textasciitilde7--9B parameter range: 
Aya Expanse 8B \citep{dang2024aya}, 
Llama 3.1 8B Instruct \citep{dubey2024llama}, 
Gemma 2 9B Instruct \citep{team2024gemma}, 
Qwen 2.5 7B Instruct \citep{yang2024qwen2}, 
EuroLLM 9B Instruct \citep{martins2024eurollm},
Aya 23 8B \citep{aryabumi2024aya},
and Salamandra 7B Instruct \citep{gonzalezagirre2025salamandratechnicalreport}.
Inference is performed using the AlpacaEval repository \citep{alpaca_eval}, with the default decoding settings: temperature 0.7, maximum tokens 2048, and models loaded in \texttt{bfloat16}.

\paragraph{Zero-Shot Results} We show our benchmark scores in \Cref{tab:xalpacaeval_zsresults}. Aya Expanse leads the table, achieving a ~60\% win rate against GPT-4o-mini for the four languages it supports (\texttt{por}, \texttt{deu}, \texttt{zho}, \texttt{hin}). While it was trained on significant synthetic data using multilingual experts \citep{dang2024aya}, it remains unclear whether its superiority stems from explicit cross-lingual tuning or implicit transfer. For other languages, Llama 3.1 and Gemma 2 yield comparable win rates ranging between 10\% and 30\%. We make two critical observations here. Firstly, except for Aya Expanse, most open LLMs trail significantly behind GPT-4o-mini in cross-lingual generation, leaving much room for improvement. Secondly, the performance strongly correlates with the resourcefulness of the language. While Aya Expanse, Llama 3.1, and Gemma achieve win rates of  40\% or higher for high-resource languages like \texttt{por}, \texttt{deu}, and \texttt{zho}, performance drops to 20-30\% for medium-resourced languages (\texttt{hun}, \texttt{lit}, \texttt{hin}) and 10\% or less for lower-resourced languages like \texttt{gle} and \texttt{mlt}. This underscores the need for scalable pipelines for creating high-quality synthetic data for lower-resourced languages, in order to achieve more consistent model performance (see Table \ref{tab:xl-alpaca-eval_full}).

\paragraph{Reason-then-Translate Results} Previous works have proposed prompting LLMs to reason first in a high-resource language (e.g., English) and then translating into the target language \citep{qin-etal-2023-cross, huang-etal-2023-languages, wang-etal-2025-large}. We call this approach ``reason-then-translate'' and report results in Table \ref{tab:rtt-results}. The outcomes are mixed: stronger multilingual models like Aya Expanse, Llama, and Gemma suffer significant performance drops. Manual inspection reveals these 7B models occasionally produce empty outputs, likely due to difficulty in following complex multi-step instructions---this aligns with prior findings which report successful results from only larger models \citep{hu-etal-2025-fine-tuning}.  In contrast, weaker LLMs like EuroLLM and Salamandra, fine-tuned on English reasoning and MT data, can leverage this two-step approach to yield some gains over their poor initial scores. Overall, these results show that inducing cross-lingual capabilities in standard multilingual LLMs may not be resolved through prompting strategies alone.
\label{sec:x-alpacaeval}

\section{The \xlinstruct{} Data Synthesis Pipeline}
\label{sec:xl-instruct}

To address this gap, we introduce the \xlinstruct{} pipeline to create cross-lingual synthetic instructions from a given seed corpus, as illustrated in \Cref{fig:xl-instruct}. We highlight two important considerations. First, unlike related work \citep{li-etal-2024-x}, we seed from English data instead of using the target language corpora directly. Given teacher LLMs are more proficient in English than in a low-resource language, we hypothesize that more high-quality, yet diverse, synthetic data could be generated in English. Machine translation is employed only in the final stages, thereby minimizing noise propagation. Second, we exclusively utilize open-weight models with permissible licenses to generate synthetic data, aligning with our objective of releasing a fully public open-source dataset. 

The \xlinstruct{} pipeline contains four stages:

\begin{enumerate}
    \item \textbf{Stage 1 Reverse Instructions:} Given a passage from our seed data, we ask a teacher LLM to generate an \textit{instruction} for which this passage would be a valid response. 

    \item \textbf{Stage 2 Refinement:} Next, we ask the teacher to reword the question and response pairs to follow four manually defined criteria.
    
    \item \textbf{Stage 3 Response Translation:} Then, we translate the refined response to the target language, using one or more translation LLMs.
    
    \item \textbf{Stage 4 Filtering:} Finally, to ensure we use the highest quality targets, we use translation quality estimation (QE) models to filter the dataset for the best translations.
\end{enumerate}
After the data is synthesized, we conduct supervised fine-tuning (SFT) on it with a range of models. We detail the minutiae in each subsection below.

\subsection{Stage 1: Question Generation}
\label{sec:stage1_generation}

First, we sample an English passage from our seed corpus, CulturaX \citep{nguyen-etal-2024-culturax}. Then, we ask a teacher LLM (Qwen 2.5 72B \citep{yang2024qwen2}) to produce an instruction \emph{for which} the sampled sentence would be a valid response. Prompting in English allows us to leverage the teacher model directly without requiring the additional fine-tuning employed previously \citep{li-etal-2024-x}. This stage thus yields a synthetic English instruction, paired with the English seed passage as a response.

\subsection{Stage 2: Refinement }
\label{sec:stage2_refinement}

Next, inspired by Self-Refine \citep{madaan2023self}, we use the teacher LLM (again, Qwen 2.5 72B) to refine the question-response pair further. Based on the most commonly occurring errors observed from manual inspection, we define four goals for the refinement process:

\begin{enumerate}
    \item \textbf{Question Self-Sufficiency:} The question should be clear and unambiguous, and should not require any additional information or context to produce the given response.
 
 \item \textbf{Response Naturalness: } The response should be `natural-sounding' as an LLM output --- in terms of fluency, neutrality, objectivity, and consistency with the tone and style of LLM-generated responses.
 
\item \textbf{Response Precision:} The response should be topically relevant, factually accurate, and should directly answer the question. This can be thought of as analogous to precision since it tries to assess how much of the information contained in the response is relevant, necessary, and true.

\item \textbf{Response Informativeness:} The response should be informative and helpful, and must contain enough justification and explanation to make it useful to an end user. This is similar to recall, as it evaluates how much of the relevant and useful information for the response is actually provided.

\end{enumerate}

We provide all four criteria and their definitions in a prompt and ask the teacher to refine the (question, response) pair. We also instruct the model to ensure the reworded response is grounded in the original one, and request it not to add any of its own knowledge---in order to avoid excessive teacher distillation and to ensure our targets are grounded in the seed data we use.

\begin{figure*}[t]
    \centering
    \begin{subfigure}[b]{0.4\linewidth}
        \centering
        \includegraphics[width=\linewidth]{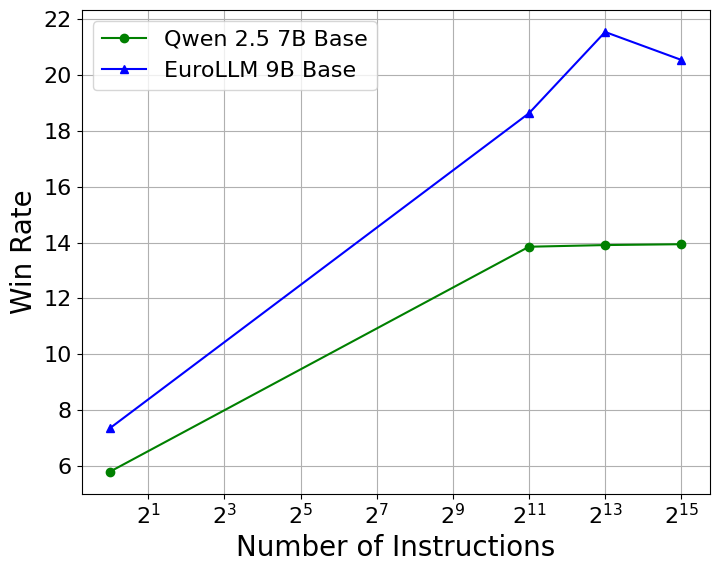}
        \caption{Base LLMs}
        \label{fig:basemodels}
    \end{subfigure}
    \hspace{8ex}
    \begin{subfigure}[b]{0.4\linewidth}
        \centering
        \includegraphics[width=\linewidth]{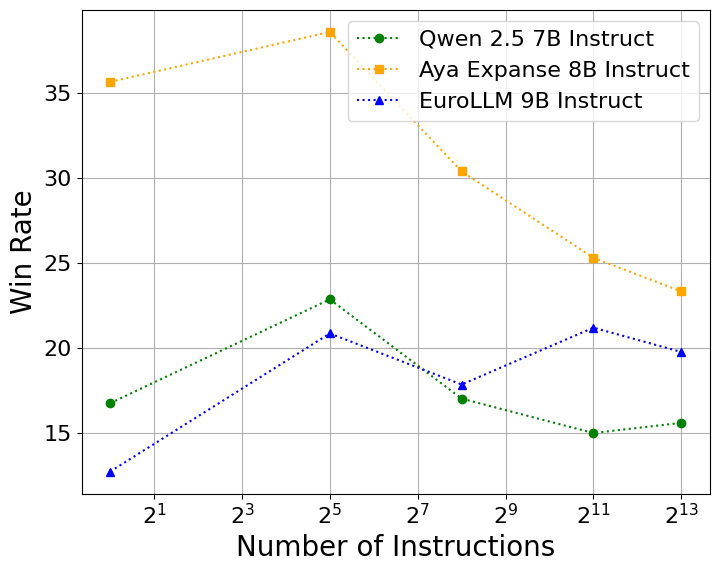}
        \caption{Instruction-Tuned LLMs}
        \label{fig:instructmodels}
    \end{subfigure}\hfill
    \caption{Performance on \xlalpacaeval{} after SFT with \xlinstruct{} data of varying sizes. Y-axis scores reflect win rates against GPT-4o-mini, averaged across 8 languages, with GPT-4o as the judge. X-axis instruction counts are shown on a log scale.}
    \label{fig:xlinstruct_xlalpacaeval}
\end{figure*}

\subsection{Stage 3: Response Translation}
\label{sec:stage3_translation}

Now, we direct our focus towards converting the English question-response pair to a cross-lingual En-X one. Creating the cross-lingual instruction itself is easy --- we simply add a prompt to ``Respond in \texttt{\{lang\}}'' where \texttt{\{lang\}} is the target language of interest.  To create the target, the English response must be machine-translated into the target language. Since document-level MT by open LLMs is currently unreliable due to limited exploration, scarce datasets, and hallucination risks, we use sentence-level translation instead. We sentence-split using Segment Any Text \citep{frohmann-etal-2024-segment} and generate translations in one of two ways:

\begin{enumerate}
    \item \textbf{Naive:} In the vanilla case, we simply prompt an LLM for the translation.
    \item \textbf{Best-of-k:} We obtain $k$ translations from $k$ different LLMs for each sentence, and choose the one with the best QE score.
\end{enumerate}

For QE, we use 
the WMT'23 CometKiwi-XL model \citep{rei2023scaling}, which obtained state-of-the-art scores in the WMT 2023 QE Shared Task \citep{blain2023findings}. For translation, we use EuroLLM 9B Instruct \citep{martins2024eurollm} in the ``naive'' case due to its strong translation capabilities, while in the ``best-of-k'' setting, we set $k=3$ and sample among EuroLLM 9B Instruct, Mistral Small 24B Instruct \citep{mistral2025small3}, and Gemma2 27B Instruct \citep{team2024gemma}. Finally, to create the translated response, we substitute each sentence in the original response with its sentence-level translation. This helps us retain formatting like paragraph separators, bullet points, etc, that is critical to response quality.

\subsection{Stage 4: Filtering}
\label{sec:stage4_filtering}

Finally, to ensure that we select high-quality targets during fine-tuning, we compute sentence-level QE scores using the WMT'23 CometKiwi-XL model by comparing each source sentence in a given response and its translation. We average these QE scores across the entire passage to obtain the final passage-level score. 
Then, we sort all responses in descending order and filter the last 20\% of the dataset -- creating a final dataset of about 32K instructions. 
We release the prompts used in each stage on Hugging Face\footnote{\url{https://huggingface.co/datasets/viyer98/xl-instruct/resolve/main/data_prompts.py}}. 

\section{Experiments on \xlalpacaeval{}: Boosting Cross-Lingual Generation}

\paragraph{Models} We conduct SFT of two base (EuroLLM 9B, Qwen2.5 7B) and three instruction-tuned (EuroLLM 9B Instruct, Qwen2.5 7B Instruct, and Aya Expanse 8B) models. We choose EuroLLM and Qwen since they relatively underperform on the \xlalpacaeval{} benchmark (Table \ref{tab:xalpacaeval_zsresults}), leaving significant scope for improvement. We also experiment with Aya Expanse since it leads the benchmark, and we are interested in seeing how much further it could be improved. Unfortunately, Aya Expanse does not have a base model released, so we are unable to experiment with it.

\paragraph{Experimental Setting} We fine-tune all models for 1 epoch using low-rank adaptation \citep[LoRA,][]{hu2022lora} with rank 8 matrices applied to query and value projections. We also tune the input and output embeddings. Training used a cosine learning rate scheduler with a peak learning rate of 1e-4 and 3\% warmup steps. We used \texttt{bf16} mixed-precision training with batch size 8, and fixed the random seed at 1 for reproducibility. All experiments were run on 4 Nvidia GeForce RTX 3090 GPUs, each with 24 GB VRAM.

\begin{table*}[t]
\centering\small
\begin{tabular}{lrrrr}
\toprule
\textbf{Model} & \textbf{Avg} & \textbf{fra} & \textbf{fin} & \textbf{tur} \\
\midrule
EuroLLM 9B & 7.80 & 9.69 & 9.78 & 3.94 \\
EuroLLM 9B Instruct & 14.08 & 19.39 & 14.05 & 8.81 \\
EuroLLM 9B XL-Instruct (best) & \textbf{20.62} & \textbf{25.80} & \textbf{22.72} & \textbf{13.33} \\
\bottomrule
\end{tabular}%
\caption{Fine-tuning with XL-Instruct yields zero-shot boosts in cross-lingual performance. Scores represent zero-shot win rates of various LLMs against GPT-4o-mini, with GPT-4o as a judge. For the XL-Instruct baseline, we use the best-performing model from Figure \ref{fig:xlinstruct_xlalpacaeval}.}
\label{tab:xli_zeroshot}
\end{table*}

\subsection{Main Results} 

In \Cref{fig:xlinstruct_xlalpacaeval}, we report \xlalpacaeval{} win rates on fine-tuning with various amounts of \xlinstruct{} data. 
We observe that for base LLMs, performance steadily improves with data scale. Qwen advances from a win rate of 5.8\% to 13.89\% against GPT-4o-mini, while EuroLLM achieves an even larger boost, going from 7.36\% to as high as 21.89\% on SFT with 8K instructions. We report language-specific scores in Table \ref{tab:xl-alpaca-eval_full} and observe that while there are consistent gains for all languages, the largest gains are on an LLM's pre-training language. Since EuroLLM includes all 8 \xlalpacaeval{} languages in its pre-training, it observes large gains per language, leading to a much better overall average score. Qwen, which chiefly supports high-resource languages like Chinese, Portuguese, and German, gains the most for these pairs but shows relatively smaller improvements for others. This suggests that while post-training with \xlinstruct{} can yield stable improvements across multiple languages, multilingual pre-training is crucial for best performance.

We also observe consistent improvements when fine-tuning instruction-tuned LLMs (\Cref{fig:instructmodels}). Unlike base LLMs, the saturation occurs sooner here---at around 2K instructions for EuroLLM-9B-Instruct and, only 32 instructions for the Qwen and Aya Expanse models! This is likely because the latter two models have also undergone Preference Optimization, and task-specific SFT at scale might lead to overfitting and deteriorated performance. EuroLLM, on the other hand, has only undergone SFT, and can therefore be trained for longer. 

Moreover, we report full results in \Cref{app:full_xl_alpacaeval_results}, where we observe consistent and major gains across all languages (Table \ref{tab:xl-alpaca-eval_full}). These results are particularly noteworthy given the modest training costs---low-rank fine-tuning with a few thousand instructions. 
Moreover, with only 32 examples, Aya Expanse achieves a win rate boost from \textasciitilde57\% to \textasciitilde65\% for its supported languages, Portuguese, German, Hindi, and Chinese (\Cref{tab:xl-alpaca-eval_full}).
Lastly, we also show in \Cref{tab:xli_zeroshot} in the Appendix how \xlinstruct{} can also boost zero-shot cross-lingual performance, i.e. even for languages \emph{not} included in SFT.

\paragraph{Zero-Shot Results} 
Moreover, we show in Table \ref{tab:xli_zeroshot} zero-shot cross-lingual performance after fine-tuning with \xlinstruct{}. We choose French (\texttt{fra}), Finnish (\texttt{fin}), and Turkish (\texttt{hind}), which are in EuroLLM's pre-training. We see huge gains in win rates, largely outperforming even the official EuroLLM 9B Instruct. This shows that even if done only for a few languages, \xlinstruct{} can result in significant transfer that improves performance in others. We hypothesize that this is likely because the model is able to learn formatting, response structure, etc. from this process, which also supports the boosts in English generation one observes in Table \ref{tab:m-alpacaeval_results}.

\subsection{Fine-Grained Evaluation} 

Beyond win rates that focus solely on pairwise comparisons, we are also interested in evaluating how well the produced cross-lingual generations improve on an absolute scale, on human-desired criteria. To achieve this, we take inspiration from recent works that define customised, task-specific metrics and use LLM-as-a-Judge for producing scores on a Likert scale, which achieves strong correlations with human ratings on the evaluation of summarization \citep{liu-etal-2023-g}, retrieval \citep{upadhyay2024umbrela}, story generation \citep{chiang-lee-2023-large}, translation \citep{kocmi-federmann-2023-large, kocmi-federmann-2023-gemba-mqm}, and open-ended generation \citep{kim2023prometheus}. In particular, \citet{kim2023prometheus} showed that using clearly defined rubrics can result in up to 0.87 Spearman correlations with human preferences for open-ended generation. Inspired by this, we propose four criteria pertinent to the task of cross-lingual generation: Objectivity, Naturalness, Informativeness, and Precision. We define detailed rubrics for each metric and provide well-defined criteria for mapping output quality to scores on a scale of 1-5. We include these rubrics in the context of a prompt, and ask GPT-4o to score cross-lingual generations of EuroLLM 9B, EuroLLM 9B Instruct, and EuroLLM 9B XL-Instruct (the best model from Figure \ref{fig:basemodels}, which is fine-tuned with LoRA on 8K examples). We provide detailed evaluation prompts and rubrics on Hugging Face\footnote{\url{https://huggingface.co/datasets/viyer98/xl-instruct/resolve/main/eval_prompts.py}}.

\begin{table*}[t]
\centering\small
\begin{tabular}{lccccc}
\toprule
\textbf{Model} & \textbf{Avg} & \textbf{Precision} & \textbf{Informativeness} & \textbf{Naturalness} & \textbf{Objectivity} \\
\midrule
EuroLLM 9B & 2.43 & 2.52 & 2.69 & 2.25 & 2.27 \\
EuroLLM 9B Instruct (official) & 3.56 & 3.68 & 3.80 & 3.54 & 3.23 \\
EuroLLM 9B XL-Instruct (our best) & \textbf{3.60} & \textbf{3.63} & \textbf{3.88} & \textbf{3.64} & \textbf{3.24} \\
\bottomrule
\end{tabular}%
\caption{Performance of EuroLLM 9B models evaluated by Precision, Informativeness, Naturalness, and Objectivity.}
\label{tab:performance_metrics}
\end{table*}

\begin{table*}[th]
\centering\small
\begin{tabular}{lrrrr}
\toprule
\textbf{Model} & \textbf{Avg} & \textbf{fin} & \textbf{hin} & \textbf{tur} \\
\midrule
EuroLLM 9B + X-Instruction (full 1M) & 9.73 & 14.35 & 7.22 & 7.63 \\
EuroLLM 9B + X-Instruction (40K) & 10.44 & 13.69 & 8.76 & 8.86 \\
EuroLLM 9B + XL-Instruct (naive, 40K) & 12.06 & 15.30 & 10.9 & 9.98 \\
EuroLLM 9B + XL-Instruct (best of 3, 40K) & \textbf{17.82} & \textbf{23.15} & \textbf{15.8} & \textbf{14.52} \\
\bottomrule
\end{tabular}
\caption{Results from EuroLLM 9B fine-tuned on data from X-Instruction \citep{li-etal-2024-x} and XL-Instruct (ours).}
\label{tab:xli_vs_xins}

\end{table*}


We list rubric-based evaluation results in \Cref{tab:performance_metrics}, which provides the macro-averaged scores across criterion and model. As expected, the raw EuroLLM 9B base model achieves the worst scores on all metrics, with the EuroLLM-9B-Instruct model performing substantially better. We note that the \xlinstruct{} model performs comparably to or marginally better than EuroLLM-9B-Instruct. This result is particularly impressive given the \xlinstruct{} baseline was trained using LoRA fine-tuning on only 8K synthetic samples, whereas the EuroLLM-9B-Instruct was fully fine-tuned on a mix of 2M human and synthetic examples. These results clearly demonstrate the effectiveness and high quality of the \xlinstruct{} dataset.

\subsection{Comparison to Previous Work}

We also compare the efficacy of \xlinstruct{} with its most similar work---the only cross-lingual open-ended generation dataset we are aware of: X-Instruction \citep{li-etal-2024-x}. 
We base all comparisons on their public data,\footnote{\url{https://huggingface.co/datasets/James-WYang/X-Instruction}} using Hindi (\texttt{hin}), Finnish (\texttt{fin}), and Turkish (\texttt{tur}), because these are supported by EuroLLM and are also available in X-Instruction. We also generate XL-Instruct data in these languages, by redoing the XL-Instruct pipeline (Section \ref{sec:xl-instruct}) from Stage 3 (Response Translation) for these languages. We LoRA fine-tune EuroLLM 9B on various X-Instruction and XL-Instruct datasets. For the former, we use both the entire 1M-sized dataset available for these languages (in total) and a 40K instructions subset, which is more comparable to our XL-Instruct baselines. For XL-Instruct, we train two baselines---one trained on ``naive'' translations (i.e., using only EuroLLM 9B Instruct) and another using a ``best-of-3'' method (refer to \Cref{sec:stage3_translation} for a detailed explanation).

We see that both XL-Instruct baselines significantly outperform X-Instruction, with our best model achieving a 70.68\% improvement over the latter---showcasing the relative superiority of our pipeline. This also suggests it might be more effective to prompt a teacher model in English due to inherently superior capabilities, and we hypothesize it might allow for greater quality and diversity in responses, as well as allow for more complex operations like refinement following specifically defined, custom criteria.
\label{sec:experiments_xl}

\section{Experiments on m-AlpacaEval: Exploring Zero-Shot Transfer}

\begin{table*}[t]
\centering\small

\begin{tabular}{lrrrrrrrrrr}
\toprule
\textbf{Model} & \textbf{Avg} & \textbf{zho} & \textbf{deu} & \textbf{hin} & \textbf{hun} & \textbf{gle} & \textbf{lit} & \textbf{mlt} & \textbf{por} & \textbf{eng} \\
\midrule
\textbf{EuroLLM 9B}  & 0.73 & 1.19  & 1.47  & 0.70  & 0.14 & 0.14  & 0.65  & 0.31  & 1.25  & 35.59 \\
\hspace*{2mm}+\xlinstruct{} (best, 8K) & 
\cellcolor{gray!35}6.10 & 
\cellcolor{gray!47}10.77 & 
\cellcolor{gray!48}11.40 & 
\cellcolor{gray!31}4.47  & 
\cellcolor{gray!27}2.53  & 
\cellcolor{gray!30}3.60  & 
\cellcolor{gray!29}3.77  & 
\cellcolor{gray!26}2.49  & 
\cellcolor{gray!44}9.76  & 
\cellcolor{gray!65}51.35 \\
\midrule
\textbf{EuroLLM 9B Instruct}  & 8.94 & 13.38 & 11.99 & 8.13  & 4.81  & 5.65  & 6.68  & 6.78  & 14.12  & 55.58 \\
\hspace*{2mm}+\xlinstruct{} (best, 8K) & 
\cellcolor{gray!39}15.55 & 
\cellcolor{gray!38}19.57 & 
\cellcolor{gray!38}18.30 & 
\cellcolor{gray!34}13.03 & 
\cellcolor{gray!36}\textbf{10.38} & 
\cellcolor{gray!44}\textbf{14.12} & 
\cellcolor{gray!49}\textbf{16.76} & 
\cellcolor{gray!41}\textbf{14.13} & 
\cellcolor{gray!31}18.11 & 
\cellcolor{gray!31}59.44 \\
\midrule
\textbf{Qwen 2.5 7B}  & 2.04 & 10.40 & 1.52  & 0.98  & 0.24  & 0.03  & 0.45  & 0.29  & 2.39  & 46.93 \\
\hspace*{2mm}+\xlinstruct{} (best, 8K) & 
\cellcolor{gray!30}5.66  & 
\cellcolor{gray!49}20.43 & 
\cellcolor{gray!43}9.53  & 
\cellcolor{gray!24}2.23  & 
\cellcolor{gray!21}0.65  & 
\cellcolor{gray!20}0.18  & 
\cellcolor{gray!23}1.60  & 
\cellcolor{gray!20}0.29  & 
\cellcolor{gray!43}10.33 & 
\cellcolor{gray!46}55.92 \\
\midrule
\textbf{Qwen 2.5 7B Instruct}  & 11.47 & 45.29 & 10.53 & 5.71  & 0.97  & 0.99  & 3.22  & 1.63  & 23.39  & 75.16 \\
\hspace*{2mm}+\xlinstruct{} (best, 32) & 
\cellcolor{gray!39}18.19 & 
\cellcolor{gray!39}52.12 & 
\cellcolor{gray!75}31.64 & 
\cellcolor{gray!27}8.34  & 
\cellcolor{gray!34}5.79  & 
\cellcolor{gray!22}1.59  & 
\cellcolor{gray!27}5.79  & 
\cellcolor{gray!21}1.83  & 
\cellcolor{gray!63}38.44 & 
\cellcolor{gray!24}76.72 \\
\midrule
\textbf{Aya Expanse 8B}  & 29.90 & 58.21 & 56.91 & 56.68 & 1.11  & 1.02  & 3.04  & 2.94  & 59.29  & 76.26 \\
\hspace*{2mm}+\xlinstruct{} (best, 32) & 
\cellcolor{gray!27}32.31 & 
\cellcolor{gray!34}63.24 & 
\cellcolor{gray!27}59.53 & 
\cellcolor{gray!39}63.22 & 
\cellcolor{gray!23}2.21  & 
\cellcolor{red!30}0.78  & 
\cellcolor{gray!28}5.85  & 
\cellcolor{gray!22}3.66  & 
\cellcolor{gray!22}60.01 & 
\cellcolor{gray!24}77.70 \\
\bottomrule
\end{tabular}
\caption{Win Rates of LLMs and their \xlinstruct{} fine-tuned counterparts on m-AlpacaEval against GPT-4o-mini, judged by GPT-4o. For each model, we choose the best cross-lingual performing baseline from Figure \ref{fig:xlinstruct_xlalpacaeval} and evaluate transfer on m-AlpacaEval. Consistent improvement across all models and pairs shows strong zero-shot transfer from cross-lingual tuning, for both multilingual and English-only generation. Best scores are bolded and cells are highlighted proportionate to performance gain.}
\label{tab:m-alpacaeval_results}
\end{table*}

Having seen task-specific improvements, we now seek to evaluate the zero-shot performance of models fine-tuned with \xlinstruct{} on multilingual and English open-ended generation, since these are arguably the more common use cases of LLMs. For this purpose, we first construct the m-AlpacaEval benchmark by machine translating the AlpacaEval test set into our 8 languages of interest, following similar efforts to create m-ArenaHard \citep{dang2024aya}. We use GPT-4o for translation of the prompts. The evaluation setup is similar to \xlalpacaeval{}, wherein GPT-4o-mini is used as the reference model and GPT-4o is used as the judge.

We present our results in Table \ref{tab:m-alpacaeval_results}, for the base and instruct versions of the LLMs from Figure \ref{fig:xlinstruct_xlalpacaeval}, alongside their best-performing \xlinstruct{}-tuned counterparts. We observe significant and consistent zero-shot transfer across all models and languages. For multilingual generation, the gains are strongest for the languages a model is pre-trained on, similar to our observations for cross-lingual generation. This is particularly evident in the Qwen and Aya models. EuroLLM Instruct, on the other hand, achieves stable performances across all languages and relatively strongest win rates for the lower-resourced languages. Interestingly, we also note consistent gains in English-only generation, despite there being no English responses on the target side! This suggests that all of these models, trained heavily on English, can learn preferred response structure and formatting from cross-lingual tuning. These results are quite encouraging, since they suggest cross-lingual fine-tuning need not come at the cost of standard ``monolingual'' generation performance---on the contrary, it can result in further boosts.

\label{sec:experiments_m}

\FloatBarrier
\section{Conclusion}
\label{sec:conclusion}

In this work, we propose data resources for advancing cross-lingual open-ended generation—loosely defined as a task in which the query and the desired (open-ended) response are in different languages. This can be viewed as a distinct yet crucial subtask of multilingual generation. While cross-lingual generation may also include more complex scenarios, such as providing context in one language while the query and response are in another (or even multiple) languages, we focus here on the simpler scenario: queries posed in English with responses required in one of eight target languages -- which includes high, medium, and low-resource EU and non-EU languages. 

With this goal in mind, we make three key contributions. First, we introduce the \xlalpacaeval{} benchmark to evaluate the current state of open LLMs, and report poor performances and significant gaps against GPT-4o-Mini. Second, we propose the \xlinstruct{} technique, and show that this synthetic data can substantially boost cross-lingual performance, both in terms of win rates and fine-grained quality metrics. Third, we show that it exhibits strong zero-shot transfer to monolingual generation, both in English and beyond. Based on these results, we strongly encourage researchers to post-train and evaluate their multilingual LLMs on our publicly released \xlsuite{}.

\section{Limitations}
\label{sec:limitations}
There has been some concern in the literature that iterative training on synthetic data could eventually lead to model collapse \citep{shumailov2024ai}. Like any other synthetic data technique, \xlinstruct{} could also share similar risks, especially since its seed data is sourced from the Web. 

We also did not perform human evaluation on the synthesized data or the model-generated outputs due to cost and time considerations. This drawback may have been partially mitigated by the rubric-based LLM judgments.


\section*{LLM Usage Statement}

AI assistants were used to aid the programming and writing process in this work. For coding, it was used to create helper functions for preprocessing, and resolve bugs. During writing, it was used to aid in constructing LaTeX tables, plot graphs, fix grammar, etc.

\section*{Contributions}
We list author contributions loosely following the CRediT author statement.\footnote{\url{https://www.elsevier.com/researcher/author/policies-and-guidelines/credit-author-statement}}

\begin{itemize}
    \item \textbf{VI}: Conceptualization (co-lead), Software (lead), Writing (lead), Methodology (lead), Formal analysis (lead), Data Curation (lead), Visualization (lead)
    \item \textbf{PC}: Conceptualization (co-lead), Writing (supporting), Supervision (supporting)
    \item \textbf{RR}: Writing (supporting), Supervision (supporting)
    \item \textbf{AB}: Writing (supporting), Supervision (lead), Project Administration (lead), Funding Acquisition (lead)
\end{itemize}

\section*{Acknowledgments}
This work has received funding from UK Research and Innovation under the UK government's Horizon Europe funding guarantee [grant numbers 10039436 and 10052546]. Vivek Iyer was supported by the Apple Scholars in AI/ML PhD fellowship. Finally, we thank EDINA team at the University of Edinburgh for their provision of OpenAI credits through the ELM API that facilitated all the experiments in this work.

We thank Wenhao Zhu for discussions during prior work \citep{iyer-etal-2024-exploring} that helped in the formulation of this idea. We also thank Simran Khanuja for her feedback on initial drafts of this work.

\bibliography{custom}


\appendix
\section{Data}

\subsection{The XL-AlpacaEval Benchmark}
\label{sec:x-alpacaeval_appendix}

Here we provide some additional details on the creation of the XL-AlpacaEval benchmark, which has 797 cross-lingual prompts in total, and currently supports 11 languages - the 8 languages used for the primary experiments in this work (Chinese, German, Hindi, Hungarian, Irish, Lithuanian, Maltese and Portuguese) and 3 additional languages (French, Finnish and Turkish) which we use for zero-shot evaluation in future sections. It is trivial to extend it to other languages -- one simply has to run a script to append cross-lingual generation instructions (Section \ref{sec:generationprompts_appendix}) to our filtered AlpacaEval test set (Section \ref{sec:manual_verification}) and such extensions are being planned as a part of future work.

\subsubsection{Manual Verification}
\label{sec:manual_verification}
\begin{table*}[t]
\centering\small
\resizebox{\textwidth}{!}{%
\begin{tabular}{@{}>{\centering\arraybackslash}m{2cm}p{0.95\textwidth}@{}}
\toprule
\textbf{Prompt ID} & \textbf{Prompt Text} \\ 
\midrule
{183} & Write a story about Anakin Skywalker encountering a Jedi who speaks and acts like a 1920s British aristocrat. \\
{200} &Write "Test" \\
{350} & I'm an English speaker trying to learn Japanese Kanji using mnemonics. Mnemonics for Kanji are created from the primitives that make them up. The Kanji for Tax has the primitives wheat and devil, so an example would be, "Taxes are like the devil taking away your hard earned wheat". Can you create a mnemonic for the Kanji meaning Wish that has the primitives clock and heart? \\
{458} & Give me a list of 5 words where the letters of the words are in alphabetical order. One example: "doors". "d" comes before "o", "o" comes before "r", and "r" comes before "s". \\
{476} & Rewrite the given text and correct grammar, spelling, and punctuation errors. If you'd told me year ago that today I would finish a marathon, I would of laughed. Your support had a huge affect on me! \\
{495} & During writing, we added an asterisk for the word that did not come to mind. You will need to provide several examples to demonstrate all the words that can be used in the sentence instead of the asterisk. \\
{635} & Correct the transcription of an excerpt containing errors. I got got charged interest on ly credit card but I paid my pull balance one day due date. I not missed a pavement year yet. Man you reverse the interest charge?  \\
{662} & You should capitalize the sentence according to the guide. Guide: Every other letter alternates between lower case and upper case.
 Sentence: A giant spider blocks your path.  \\
{663} &  Create alliterations by finding synonyms for words in the given sentence. David wears a hat everyday.  \\
{714} &  Rewrite the text and correct the spelling errors. It solves problems comon and uniqe to every team.  \\
\bottomrule
\end{tabular}%
}
\caption{Culturally specific prompts removed from the AlpacaEval dataset.}
\label{tab:removed_prompts}
\end{table*}

\begin{table*}[t]
\centering\small
\setlength{\tabcolsep}{24pt} 
\begin{tabular}{@{}ll@{}}
\toprule
\textbf{Prompts} & \\
\midrule
Answer in \{\} language & Output an answer in \{\} language \\
Generate your answer in \{\} language & Respond in \{\} language \\
Produce an answer in \{\} language & Please write in \{\} language \\
\bottomrule
\end{tabular}
\caption{Cross-Lingual Generation Instructions}
\label{tab:xlg-instructions}
\end{table*}

Before creating our cross-lingual benchmark, we conduct a rigorous stage of manual verification to ensure that the prompts are suitable for answering cross-lingually. In Table \ref{tab:removed_prompts}, we show the prompts we removed from AlpacaEval that were too culturally specific (for instance, prompt 183) or tailored towards eliciting an English response (prompts 350 and 714). In the latter, we felt mandating a non-English response might make evaluating a ``correct" response challenging. In other cases where the prompt simply requested a response in English, we replaced with a generic templated variable \texttt{\{language\}} for downstream substitution with the name of the desired target language. This leaves us with a total of 797 prompts. It is important to note that as far as possible, we tried to keep complex multi-step, multilingual prompts in our evaluation set, and only removed cases that were clearly invalid -- in keeping with the goal of this work to build robust cross-lingual models.

\subsubsection{Generation prompts}
\label{sec:generationprompts_appendix}

Next, we randomly sample prompts from a list of cross-lingual generation instructions (given in Table \ref{tab:xlg-instructions}), and append it to each prompt in the filtered test set from the previous stage. To add further diversity to the instructions in the benchmark, we remove the word ``language" from the prompts given in Table \ref{tab:xlg-instructions} -- thus converting ``Answer in German language" to ``Answer in German". This leads to the creation of the final XL-AlpacaEval benchmark.

\subsection{License}

The \xlalpacaeval{} dataset, which is derived from the AlpacaEval dataset, is released under a CC-by-NC 4.0 license, following its predecessor. This means the dataset is primarily intended for use in non-commerical (research) contexts. In contrast, the \xlinstruct{} dataset, which is provided as a training dataset, is derived from the CulturaX corpus -- which in turn sources from the mC4 \citep{xue-etal-2021-mt5} and OSCAR \citep{ortiz-suarez-etal-2020-monolingual}. mC4 is released under an ODC-BY license, and OSCAR is released under CC0 no rights reserved. Hence, \xlinstruct{} can be used in both commercial and research contexts, as long as the corresponding licenses are respected.

\section{Additional Experiments and Results}

\subsection{XL-AlpacaEval Results}\label{app:full_xl_alpacaeval_results}

\begin{table*}[t]
\centering\small
\begin{tabular}{lrrrrrrrrr}
\toprule
\textbf{Model} & \textbf{Avg} & \textbf{zho} & \textbf{deu} & \textbf{hin} & \textbf{hun} & \textbf{gle} & \textbf{lit} & \textbf{mlt} & \textbf{por} \\
\midrule
\textbf{EuroLLM 9B} & 7.36 & 8.97 & 9.96 & 4.49 & 4.13 & 6.09 & 9.94 & 4.66 & 10.61 \\
\hspace*{2mm}+2K instructions & 18.63 & 18.77 & 23.65 & 13.22 & 13.70 & 16.03 & 25.48 & 14.75 & 23.47 \\
\hspace*{2mm}+8K instructions & \textbf{21.54} & 20.98 & \textbf{26.76} & \textbf{16.26} & \textbf{17.27} & 20.99 & \textbf{28.52 }& \textbf{15.72} & \textbf{25.81} \\
\hspace*{2mm}+32K instructions & 20.54 & \textbf{21.24} & 24.07 & 15.26 & 17.11 & \textbf{21.08} & 28.09 & 15.64 & 21.79 \\
\midrule
\textbf{EuroLLM 9B Instruct} & 12.70 & 14.82 & 16.49 & 8.23 & 8.66 & 9.37 & 16.57 & 8.51 & 18.94 \\
\hspace*{2mm}+32 instructions & 20.84 & 23.52 & 22.96 & 13.10 & 17.37 & 17.10 & 25.61 & 21.30 & 25.79 \\
\hspace*{2mm}+256 instructions & 17.83 & 21.13 & 21.73 & 12.90 & 14.05 & 13.25 & 21.13 & 15.32 & 23.11 \\
\hspace*{2mm}+2K instructions & \textbf{21.18} & \textbf{23.62} & \textbf{24.39} & 14.49 & \textbf{16.63} & 20.17 & \textbf{27.87} & \textbf{18.02} & \textbf{24.22} \\
\hspace*{2mm}+8K instructions & 19.75 & 23.10 & 22.65 & \textbf{14.50} & 14.97 & \textbf{20.55} & 26.92 & 15.34 & 19.96 \\
\midrule
\textbf{Qwen 2.5 7B} & 5.80 & 12.62 & 6.36 & 3.40 & 2.73 & 4.50 & 4.33 & 2.62 & 9.82 \\
\hspace*{2mm}+2K instructions & 13.85 & 33.64 & 18.37 & 6.67 & \textbf{6.50} & 5.00 & 10.73 & \textbf{3.63} & 26.22 \\
\hspace*{2mm}+8K instructions & 13.91 & 34.22 & \textbf{19.80} & 6.61 & 6.28 & 3.92 & 10.22 & 3.07 & \textbf{27.13} \\
\hspace*{2mm}+32K instructions & \textbf{13.94} & \textbf{34.29} & 18.88 & \textbf{6.88} & 5.72 & \textbf{5.44} & \textbf{10.77} & 3.36 & 26.18 \\
\midrule
\textbf{Qwen 2.5 7B Instruct} & 16.73 & 44.63 & 16.35 & 9.59 & 6.82 & 7.17 & 14.68 & 3.69 & 30.88 \\
\hspace*{2mm}+32 instructions & \textbf{22.85} & \textbf{50.16} & \textbf{31.66} & \textbf{12.36} & \textbf{12.52} & \textbf{8.66} & \textbf{19.40} & \textbf{4.91} & \textbf{43.10} \\
\hspace*{2mm}+256 instructions & 17.00 & 38.04 & 22.45 & 9.46 & 7.85 & 5.39 & 15.86 & 4.02 & 32.92 \\
\hspace*{2mm}+2K instructions & 14.97 & 36.17 & 18.95 & 8.44 & 7.14 & 5.06 & 12.02 & 3.02 & 28.92 \\
\hspace*{2mm}+8K instructions & 15.57 & 42.74 & 18.85 & 8.32 & 6.54 & 4.41 & 11.99 & 3.49 & 28.19 \\
\midrule
\textbf{Aya Expanse 8B} & 35.67 & 57.22 & 60.27 & 56.99 & 8.62 & 10.43 & 19.54 & 9.51 & 62.75 \\
\hspace*{2mm}+32 instructions & \textbf{38.61} & \textbf{64.08} & \textbf{65.07} & \textbf{59.76} & \textbf{10.71} & \textbf{11.72} & \textbf{21.57} & \textbf{10.70} & \textbf{65.28} \\
\hspace*{2mm}+256 instructions & 30.39 & 55.65 & 52.93 & 44.50 & 6.51 & 6.45 & 17.90 & 6.07 & 53.10 \\
\hspace*{2mm}+2K instructions & 25.30 & 41.84 & 46.43 & 37.03 & 6.77 & 4.55 & 15.94 & 3.75 & 46.12 \\
\hspace*{2mm}+8K instructions & 23.32 & 43.16 & 42.23 & 28.19 & 5.44 & 6.00 & 15.94 & 4.91 & 40.72 \\
\bottomrule
\end{tabular}
\caption{Full language-wise Win Rates against GPT-4o-mini on XL-AlpacaEval, after LoRA fine-tuning on varying sizes of XL-Instruct data on different LLMs. GPT-4o is the judge. The best scores per model are highlighted in bold.}
\label{tab:xl-alpaca-eval_full}
\end{table*}

\paragraph{Full Results} In Table \ref{tab:xl-alpaca-eval_full}, we show the complete language-wise results for each base and instruct model we tuned on varying sizes of XL-Instruct data. Models like EuroLLM and Qwen continue improving until 8K-32K instructions, with gains diminishing in the last 24K instructions. This is likely because we sort the instructions in order of translation quality, and sample them accordingly, reducing the gains. It is possible that improving the translation quality further could result in larger gains. For preference-optimized (PO'ed) instruction-tuned models, performance saturates at 32 instructions, and 2K instructions with non-PO'ed models like EuroLLM 9B Instruct. The largest gains across all models are consistently for the languages included during pretraining -- for instance, Qwen 7B improves on Chinese win rates from 12.62 to 34.29 and in Portuguese from 9.82 to 27.13, suggesting the criticality of this stage in building multilingual LLMs.

\begin{table*}[t]
\centering\small
\begin{tabular}{lrrrrrrrrr}
\toprule
\textbf{Model} & \textbf{Avg} & \textbf{zho} & \textbf{deu} & \textbf{hin} & \textbf{hun} & \textbf{gle} & \textbf{lit} & \textbf{mlt} & \textbf{por} \\
\midrule
\textbf{EuroLLM 9B} & 7.36 & 8.97 & 9.96 & 4.49 & 4.13 & 6.09 & 9.94 & 4.66 & 10.61 \\
\hspace*{2mm}+8K instructions (random) & 22.69 & 22.66 & 25.71 & 15.59 & 18.45 & 22.40 & 29.52 & 20.50 & 26.72 \\
\hspace*{2mm}+8K instructions (naive) & 21.17 & 20.26 & 24.63 & 15.53 & 16.68 & 21.96 & 28.33 & 18.24 & 23.75 \\
\hspace*{2mm}+8K instructions (best of 3) & 21.54 & 20.98 & 26.76 & 16.26 & 17.27 & 20.99 & 28.52 & 15.72 & 25.81 \\
\midrule
\textbf{EuroLLM 9B Instruct} & 12.70 & 14.82 & 16.49 & 8.23 & 8.66 & 9.37 & 16.57 & 8.51 & 18.94 \\
\hspace*{2mm}+32 instructions (random) & 18.49 & 22.21 & 22.69 & 12.70 & 15.18 & 15.35 & 21.87 & 14.12 & 23.79 \\
\hspace*{2mm}+32 instructions (naive) & 18.55 & 22.20 & 20.16 & 12.18 & 13.70 & 15.14 & 23.64 & 17.55 & 23.84 \\
\hspace*{2mm}+32 instructions (best of 3) & 20.84 & 23.52 & 22.96 & 13.10 & 17.37 & 17.10 & 25.61 & 21.30 & 25.79 \\
\bottomrule
\end{tabular}%
\caption{Ablations of the strategy for selecting response translations for the EuroLLM 9B and EuroLLM 9B Instruct models.}
\label{tab:ablations}
\end{table*}

\subsection{Ablations}

Lastly, we conduct an ablation to verify the importance of the translation selection strategy. Given the cross-lingual part of the dataset mainly comes from Machine Translations, and translations can be quite noisy, we experiment with 2 MT techniques, ``naive'' and ``best-of-3'' responses. We also include a ``random'' sampling strategy, where random responses are chosen for subsampling, regardless of MT quality. We fine-tune the EuroLLM 9B and EuroLLM 9B Instruct models using 8K and 32 instructions respectively, which are respectively the optimal SFT data sizes for each model (check Figure \ref{fig:xlinstruct_xlalpacaeval}). 

For the instruct model, ``best of 3'' introduces significant improvements over naive or random sampling strategies, taking the average win rate from 18.55 to 20.84. This is likely because at the tiny scale of 32 instructions, target response quality matters hugely and significantly impacts performance. For EuroLLM 9B, which is fine-tuned on 8K instructions, performance still improves for most languages with the best-of-3 technique. The only cases where it drops are for the least-resourced languages like Irish and Maltese, which makes the average score much lower. It is possible the CometKiwi model we use for Quality Estimation is not very well-suited for such low-resource languages. As a result, we hypothesize that best-of-3 might sometimes end up choosing a worse translation than the naive method -- which uses EuroLLM, a model known to have strong MT capabilities for all these languages.

\label{sec:appendix}

\end{document}